\documentclass[11pt,a4paper]{article}
\usepackage[margin=2.5cm]{geometry}
\usepackage[T1]{fontenc}
\usepackage[utf8]{inputenc}
\usepackage{lmodern,microtype,amsmath,amssymb,booktabs,array}
\usepackage[numbers,sort&compress]{natbib}
\usepackage[hidelinks]{hyperref}
\usepackage{url,graphicx,placeins,pdflscape}
\hypersetup{pdftitle={Local subspace encoding for spectral prediction: from TabPFN to operator-adaptive calibration},
pdfauthor={Gregory Beurier, Robin Reiter, Denis Cornet, Lauriane Rouan},
pdfkeywords={NIR, LVSE, GCU, TabPFN, RamanPFN, PLS, Ridge, AOM, nirs4all}}
\newcommand{\TwoFiveMeanGain}{10.22}
\newcommand{\TwoFiveMedianGain}{4.99}
\newcommand{\TwoFiveWins}{47}
\newcommand{\ThreeMeanGain}{12.80}
\newcommand{\ThreeMedianGain}{4.74}
\newcommand{\ThreeWins}{49}
\newcommand{\ThreeFiveMeanGain}{14.52}
\newcommand{\ThreeFiveMedianGain}{7.82}
\newcommand{\ThreeFiveWins}{53}
\newcommand{\PLSMeanGain}{5.45}
\newcommand{\PLSMedianGain}{4.75}
\newcommand{\PLSWins}{47}
\newcommand{\RidgeMeanGain}{0.57}
\newcommand{\RidgeMedianGain}{1.89}
\newcommand{\RidgeWins}{36}
\newcommand{\TwoFiveFOneGain}{2.22}
\newcommand{\TwoFiveNIRFOneGain}{2.71}
\newcommand{\ThreeFOneGain}{2.56}
\newcommand{\ThreeNIRFOneGain}{2.52}
\newcommand{\ThreeFiveFOneGain}{3.81}
\newcommand{\ThreeFiveNIRFOneGain}{3.77}
\newcommand{\AOMPLSLVSESelections}{63}
\newcommand{\AOMRidgeLVSESelections}{88}

\title{Local subspace encoding for spectral prediction:\\
from TabPFN to operator-adaptive calibration}
\author{Gr\'egory Beurier\textsuperscript{1,2}\thanks{Corresponding author: \texttt{gregory.beurier@cirad.fr}.}
\quad Robin Reiter\textsuperscript{1,2}\\
Denis Cornet\textsuperscript{1,2}\quad Lauriane Rouan\textsuperscript{1,2}}
\date{\small\textsuperscript{1}CIRAD, UMR AGAP Institut, F-34398 Montpellier, France\\
\textsuperscript{2}UMR AGAP Institut, Univ Montpellier, CIRAD, INRAE, Institut Agro, Montpellier, France\\[0.6em]
1 October 2026}

\begin{document}
\maketitle
\begin{abstract}
TabPFN has already shown useful predictive performance on near-infrared spectra.
RamanPFN subsequently proposed two spectral representations for tabular inference:
Global Compositional Unmixing (GCU) and Local Vibrational Subspace Encoding (LVSE).
We assess whether these representations also help on datasets assembled through
the nirs4all ecosystem, and whether their value extends to classical calibration.
We reanalyse completed experiments from a public spectral dataset catalogue.
On a post-hoc subset of 61 matched numerical prediction tasks, after excluding
ten class-code, metadata or semantically ambiguous targets, LVSE reduces RMSE relative to raw-spectrum TabPFN by
\TwoFiveMeanGain\%, \ThreeMeanGain\% and \ThreeFiveMeanGain\% on average for
checkpoints 2.5, 3 and 3.5, respectively, with median reductions of
\TwoFiveMedianGain\%, \ThreeMedianGain\% and \ThreeFiveMedianGain\%.
Across 76 classification datasets, weighted F1 increases by
\TwoFiveFOneGain, \ThreeFOneGain\ and \ThreeFiveFOneGain\ percentage points.
The 2.5 classification procedure uses output coding on 15 datasets exceeding
the checkpoint's native class limit.
GCU alone is less reliable in this implementation. LVSE also improves PLS and
Ridge in many cases, with median RMSE reductions of \PLSMedianGain\% and
\RidgeMedianGain\%. Adding LVSE candidates to an operator-adaptive model is
feasible, but a separate 35-task panel derived from the public AOM benchmark
shows only small median gains. These results motivate evaluating LVSE alongside
existing representation choices. They do not establish that a selector choosing
between raw and LVSE inputs will improve every task.
\end{abstract}

\section{Introduction}
Near-infrared (NIR) calibration depends on how spectra are represented before a
predictive model is fitted. Adjacent channels are strongly correlated, and
normalization, derivatives and latent projections can change which information
the model uses. A useful preprocessing therefore needs to be evaluated together
with the predictor, on training data, rather than selected from its performance
on a held-out test \citep{rinnan2009,beurier2026aom}.

TabPFN offers a different predictive component: a pretrained tabular model uses
labelled examples from the new task as context, without task-specific updating
of its neural weights \citep{hollmann2025}. Our earlier NIR study evaluated raw
and preprocessing-optimized TabPFN against chemometric and machine-learning
baselines, and found that tabular foundation models can complement established
calibration workflows \citep{reiter2026}. Pan et al. then introduced RamanPFN,
which places global nonnegative coordinates and local spectral subspaces in front
of TabPFN \citep{pan2026}. Their contribution motivates the present question:
does the benefit of a spectral representation transfer beyond the Raman setting,
including data collections used in NIR research?

We study this question using experiments on the public
\texttt{nirs4all-datasets} catalogue, available at
\url{https://datasets.nirs4all.org}. The objective is a practical evaluation,
not a new spectral foundation model. We first compare raw spectra, GCU, LVSE and
their concatenation with several TabPFN checkpoints. We then apply the same
representations to PLS and Ridge. Finally, we examine whether LVSE is a useful
addition to the candidate bank of adaptive operator modelling (AOM), and compare
the extended models with TabPFN on a panel derived from the publicly described
AOM benchmark \citep{beurier2026aom}.
This provides one experimental thread from a Raman-inspired representation to
spectral tabular inference and classical calibration, while keeping the distinct
validation protocols visible.

\begin{figure}[htbp]\centering
\includegraphics[width=\linewidth]{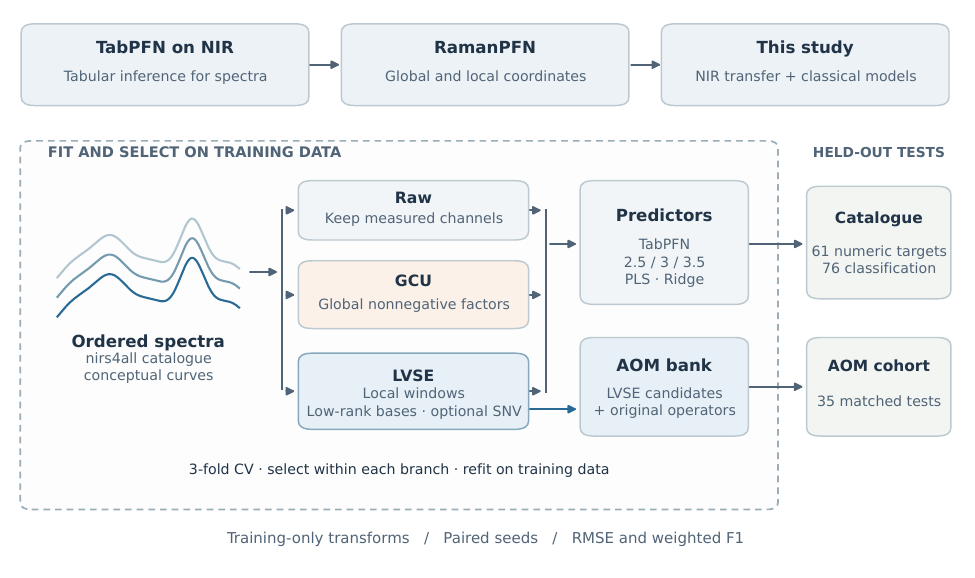}
\caption{From spectral representation to adaptive calibration. The motivating
sequence is TabPFN on NIR, RamanPFN's global/local coordinates, then their reuse
with TabPFN and classical calibration. Transformations and selection are fitted
on training data before held-out evaluation. The catalogue and AOM-derived panels
use separate protocols. Curves are conceptual, not measured spectra.}
\label{fig:overview}
\end{figure}

Figure~\ref{fig:overview} gives the overall logic. A spectral representation is
a reusable component between the measured spectrum and the predictor. The first
question is whether that component helps with a fixed predictor family. The
second is whether it adds value to a model that already selects its own
preprocessing. These questions require separate comparisons.

\FloatBarrier

\section{Materials and methods}
\subsection{Datasets and experimental scope}
The catalogue snapshot contains 164 entries, assigned by its metadata to
97 NIR, five mid-infrared (MIR), 58 \emph{other}, three Raman and one mixed. These catalogue
entries are not all successful supervised benchmarks. The analysis uses only
saved valid scores, excludes the Raman and mixed entries, and retains the
campaign's explicit exclusions of the large Kellogg and two LUCAS soil entries.
The full exploratory panel comprises 71 catalogue-declared numerical prediction
datasets and 76 classification datasets, with scores on 107 distinct identifiers.
The classification metadata assigns 55 to NIR, two to MIR and 19 to other entries.
A dataset can contribute to both regression and classification. Consequently,
these counts must not be added to infer a count of independent datasets.

The TabPFN comparison uses one representative spectral source--target pair
per available dataset and task type, with paired outer seeds 0, 1 and 2.
Completed pilot representatives are retained. Additional representatives are
chosen by greatest available archived seed coverage, then smallest declared
rows $\times$ channels, with deterministic source/target tie-breaks. The frozen
manifest records the exact keys.
The original classical comparison covers 966 numerical source--target pairs
per model, across the same 71 identifiers. We retain that all-target comparison
in the appendices and explicitly match the primary classical analysis to the
TabPFN representatives below. The catalogue is only partly scored. Some entries
are alternate spectral views of related source collections, so even equal
weighting of catalogue identifiers does not make every identifier an
independent scientific cohort. Supplementary Figure~\ref{fig:catalogcomplete}
shows the complete evaluated panels: 71 numerical and 76 classification
identifiers, with all 31 and 18 recorded procedures, respectively.

Catalogue labels are not a certified modality or target-type audit. For example,
the Coffee and Strawberry origins are FTIR classification datasets, although
their class codes were also scored as numerical targets in this panel
\citep{coffeedata,strawberrydata}. Several soil entries also originate from MIR
collections. We therefore do not describe the entire numerical panel as NIR
chemical calibration.

For the primary numerical analysis, we exclude ten representatives identified
as nominal codes, metadata or semantically ambiguous targets: coffee/meat species, strawberry authenticity,
the two ambiguous malaria targets named x, easting, day of year in two entries, and
individual identifiers in two entries. This post-hoc rule is based on target
meaning, independently of observed gain. It leaves 61 representatives. PLS,
Ridge and all three TabPFN checkpoints are compared on their exact same
source--target--seed keys. This subset does not constitute a complete assay or
modality audit. The full 71-entry results remain visible in
Appendices~\ref{app:numbers} and \ref{app:catalogue}; the excluded keys accompany
the score data.

Spectra and labels are joined by sample identifier. Missing values, nonfinite
spectra and target sentinels are removed. For classification, classes with fewer
than nine distinct recorded sample-ID groups are excluded; at least two supported
classes must remain. Repeated measurements of one sample stay in the same split.
Categorical targets are those declared by the catalogue, including species,
material and identifier-like labels. The broad classification panel therefore
measures predictive performance on these declared labels, not solely chemical
assays or clinically meaningful classes. The metadata-labelled NIR stratum is
reported separately, without claiming that all its modality labels are verified.

\subsection{Validation and model selection}
The catalogue representation study uses an 80/20 train/test split repeated with
three outer seeds. Classification splits are stratified, including at the sample
group level when repeated scans are present. Three inner folds also preserve
repeated recorded sample-ID groups: GroupKFold for numerical targets and
StratifiedGroupKFold for classification, with ordinary KFold/StratifiedKFold
otherwise. Selection uses pooled out-of-fold RMSE or weighted F1, rather than
an average of fold scores. GCU/LVSE are fitted anew inside each
fold; the selected configuration is then refitted on the full training
partition and evaluated on the untouched test. Representation fitting uses at
most 5,000 randomly sampled training spectra. TabPFN context fitting uses at
most 10,000 rows, with stratified sampling for classification. Classical outer
training is capped at 10,000 rows before inner CV. No catalogue context/estimator
cap binds in the retained scores, although the representation cap does bind.
These subsampling steps are row-wise. This remains a within-dataset evaluation: it is not a
leave-instrument-out or leave-campaign-out transfer study.

PLS selects among 2, 5 and 10 components, clipped to feasible dimensions, with
\texttt{scale=False}. All classical branches, including raw, first apply a
training-fit StandardScaler to their coordinates. Ridge selects its regularization coefficient from
$\{0.01,0.1,1,10,100\}$. For each model, each representation branch is selected
using its own inner validation predictions. The raw-spectrum comparison also
selects its model parameter on training data. Thus the classical results compare
complete branch-specific procedures; they do not isolate compression at a
fixed downstream parameter.

TabPFN checkpoints 2.5, 3 and 3.5 \citep{tabpfn25,tabpfn3,tabpfn35} are evaluated through \texttt{tabpfn} 9.0.0
with eight inference estimators, automatic estimator scaling, fixed random
seeds and low-memory inference. Each checkpoint has its own training-selected
representations in the catalogue version study. No TabPFN weights are
fine-tuned and no joint pretraining on the catalogue is performed.
These checkpoint comparisons are paired within version; they do not prove that
one model version dominates another in every application.

\subsection{Local and global representations}
The names GCU and LVSE follow RamanPFN \citep{pan2026}. We evaluate an independent
implementation of these ideas, rather than a reproduction of the authors' official implementation.
The definitions below describe the transformations used to produce our saved
scores.

LVSE divides the ordered channel axis into contiguous, nonoverlapping windows.
The candidate bank uses windows of 64 or 128 channels, retains two or four
singular modes per window, and includes branches with and without standard
normal variate (SNV) normalization. Within a training fold, channels are
standardized using their training means and standard deviations. If
$\widetilde X_b$ denotes a standardized block, a truncated singular value
decomposition supplies a local basis $V_b$. A query spectrum $x$ is encoded as
\begin{equation}
 z_{\mathrm{LVSE}}(x)=\bigoplus_b
 \left[\frac{p(x)_b-\mu_b}{\sigma_b}-c_b\right]V_b,
 \label{eq:lvse}
\end{equation}
where $p$ is either identity or SNV, $c_b$ is the training block centre and
$\oplus$ denotes concatenation. Bases and scaling parameters are training-only.
Window widths are measured in channels, not in nanometres or wavenumbers;
the same width therefore has different physical coverage across instruments.

GCU uses a nonnegative matrix factorization (NMF) across the full spectrum.
For each training fold, the implementation subtracts the channel-wise training
minimum, divides by a training scale and clips negative values to zero. NMF ranks
8, 16 and 32 and iteration limits 20 and 60 form six TabPFN candidates; classical
models use the same ranks with 60 iterations only. Query spectra
are mapped using the fitted factorization. These coordinates are statistical
factors; we do not claim that they identify chemical components. The short
iteration limits sometimes generated convergence warnings. This limitation is
part of the evaluated GCU procedure and matters when interpreting its failures.

In the catalogue version study, GCU+LVSE concatenates the two independently
selected coordinate sets. This should be distinguished from the initial TabPFN-3
evaluation's prediction-level combination, also recorded under a GCU/LVSE label.
An additional prediction combination applies a signed correction to a raw
TabPFN anchor. Eight predetermined GCU/LVSE pairs form correction candidates,
with signed weights 1, 0.75 or 0.5; a correction scale clipped to $[0,1]$ is
estimated from training out-of-fold residuals. We call this variant
\emph{raw plus signed correction}. It uses the same pretrained model weights.

\subsection{Completing multiclass evaluation}
The 2.5 classifier checkpoint cannot natively fit more than ten classes.
Rather than discard these datasets or merge labels, a separate completed
campaign applied the official \texttt{ManyClassClassifier} output-coding wrapper
from \texttt{tabpfn-extensions} 0.6.3 \citep{priorlabs2026}, with alphabet size 10,
redundancy parameter 4 and the same outer splits. It supplies 180 valid scores
for the four input variants, three seeds and 15 datasets with 11--46 classes.
The other 61 classification datasets use the native 2.5 checkpoint. The aggregate
2.5 results therefore describe a conditional procedure: native inference when
supported, output coding otherwise. Raw and transformed inputs use the same
procedure within each dataset. Native and wrapped results are separated in
Appendix~\ref{app:manyclass}.

\subsection{LVSE candidates in operator-adaptive calibration}
AOM selects preprocessing operators jointly with PLS or Ridge calibration
\citep{beurier2026aom}. For a fixed linear channel operator $A$, row spectra
transform as $XA^\mathsf{T}$, and a linear predictor can be returned to the
original wavelengths. A fitted LVSE map without SNV is affine in a new spectrum:
once its training basis is fixed, a coefficient vector in encoded space can
likewise be mapped back to channel space. The important boundary is the fitting
stage. The local bases must be learned separately inside each validation fold;
fitting them once on all calibration data before inner cross-validation would
leak validation information. SNV branches also retain their sample-dependent
normalization at inference and are not strict fixed linear operators.

The saved extension study adds LVSE candidates to AOM's operator selection and
compares them with retained original-bank AOM runs on 35 regression tasks from
15 source families in the benchmark described by Beurier et al.
\citep{beurier2026aom}. This is the available 35-task panel with archived
comparison runs and matched test partitions; it is distinct from the cited
preprint's strict headline denominator of 32. The retained original-bank engine
scores are not transcriptions of the published arXiv score table. Input files,
partitions, fold identifiers and implementation hashes are
preserved by that campaign. Direct depth-one AOM-PLS and AOM-Ridge have three
seeds; deeper chain and FastAOM configurations use the archived seed-zero
protocol. The fixed LVSE study includes the original window/mode family and
additional window/rank/overlap candidates. The separate adaptive search uses window
sizes 16, 32, 64 and 128, two or four local components, and a shortlist of four
candidates over identity and Savitzky--Golay parent branches.

The retained 35-task inner partitions are row-wise shuffled three-fold KFold
indices. They are reused across comparisons, but repeated-scan group preservation
is not established for this panel. The initial TabPFN evaluation uses package
8.0.8 and its recorded default model, labelled version 3 from that installation's
default. The base checkpoint was not explicitly pinned and no archived weight
hash is available. The later checkpoint evaluations use package 9.0.0 and frozen
training-selected representations and combination weights. Both TabPFN
procedures cap representation fitting at 5,000 rows and context fitting at
10,000. Both caps bind for the 39,225-row training partition of the large leaf
mass-per-area task; AOM uses its complete training partition. This is a
material budget difference, not a controlled equal-budget comparison.

This campaign compares an archived engine with an updated engine plus an expanded
bank. The update also changes a numerical PLS fallback. Accordingly, its score
differences are \emph{implementation-and-bank differences}, not a clean causal
LVSE ablation. Historical runtimes came from separate campaigns on shared CPUs
and are descriptive rather than controlled speed comparisons.

\subsection{Metrics and aggregation}
Regression uses held-out RMSE. For representation $r$ and raw baseline $0$ on
the same task and seed, the reported gain is
\begin{equation}
 g=100\left(1-\frac{\operatorname{RMSE}_{r}}
                         {\operatorname{RMSE}_{0}}\right).
\end{equation}
Classification uses weighted F1 and reports
$g=100(F1_r-F1_0)$ in percentage points (pp). Positive values indicate
improvement in both cases. We average paired gains over seeds within a
source--target task and then over tasks within a catalogue dataset, before
reporting the mean, median and wins across datasets with equal dataset weights.
Only valid paired keys are used. Wins and losses use a numerical tie tolerance
of $10^{-8}$ in gain units. These are descriptive summaries; we do not treat
three seeds as three independent datasets or claim statistical significance
from the win counts.

The AOM-derived panel retains its own aggregation: task scores are averaged over
the archived available seeds before task-level RMSEP ratios and gains are
summarized. Common-reference ratios use AOM-PLS depth one on the identical task
test. A median of ratios is not the ratio of two medians, and gains against
different references should not be subtracted to infer a new comparison.

\section{Results}
\subsection{LVSE improves TabPFN on matched numerical tasks}
\begin{figure}[htbp]\centering
\includegraphics[width=\linewidth]{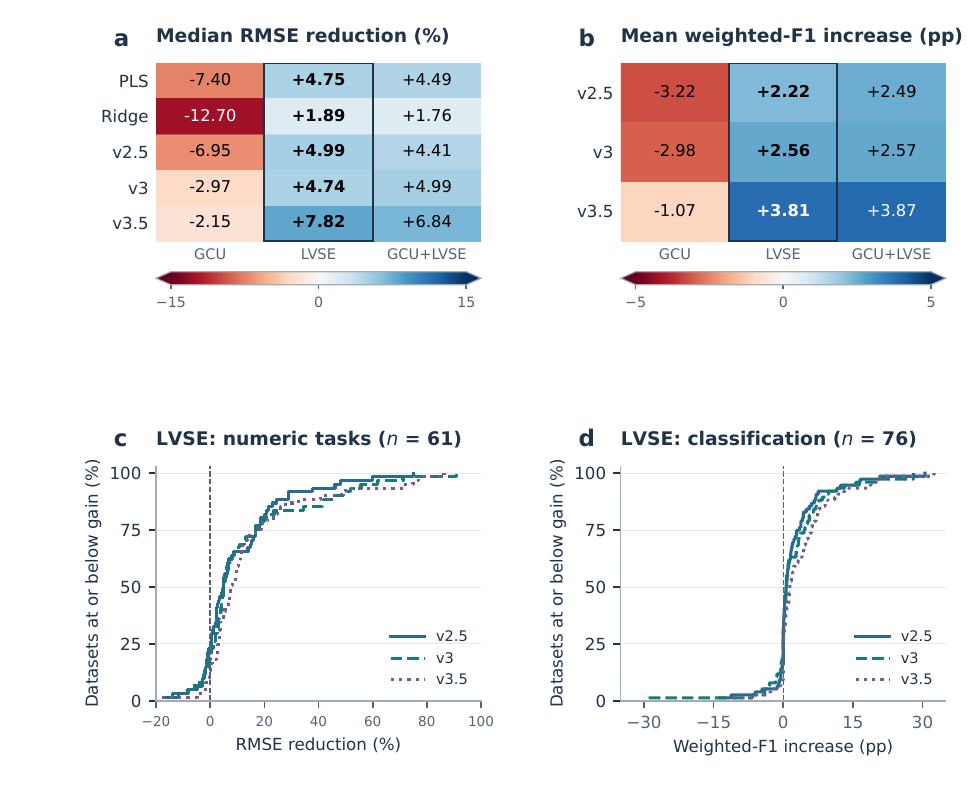}
\caption{LVSE transfers across predictors, with variable benefits across datasets.
(a) Median paired RMSE reductions on 61 numerical representatives matched across
all five predictors after the declared post-hoc exclusions; (b) mean weighted-F1
increases on 76 classification datasets. Signed cell values provide the numerical
comparison independently of colour. (c,d) Full cumulative distributions of
LVSE-versus-raw dataset gains for three checkpoints. Positive gains indicate
improvement; the dashed vertical line marks no change. Regression curves retain
all 61 gains. Dataset gains
average three paired seeds, with equal dataset weights. The 2.5 classifier uses
ManyClass on 15 datasets. These curves are descriptive distributions.}
\label{fig:representations}
\end{figure}

Figure~\ref{fig:representations}a,c shows a benefit from LVSE in all three
checkpoint versions. Mean RMSE reductions are \TwoFiveMeanGain\%,
\ThreeMeanGain\% and \ThreeFiveMeanGain\%, while the corresponding medians are
\TwoFiveMedianGain\%, \ThreeMedianGain\% and \ThreeFiveMedianGain\%.
LVSE wins on \TwoFiveWins/61, \ThreeWins/61 and \ThreeFiveWins/61 representatives,
respectively. The improvement is therefore not restricted to a few large tasks,
although adverse cases remain.
The cumulative distributions show where those gains occur: most datasets lie
on the improving side of zero, while adverse cases remain.
The figure retains all observed LVSE gains. Detailed means, medians and win counts are
reported in Table~\ref{tab:matchednumeric} in Appendix~\ref{app:numbers}.
These are the post-hoc numerical-target results. The full declared-numeric
panel has a different mean, particularly for checkpoint 3: class-code RMSE
cases contributed extreme losses. Such cases are not failures of concentration
calibration and are kept separately in Table~\ref{tab:regression}.

GCU alone has negative mean and median gains for each version. Its large negative
means, compared with more moderate medians, indicate some severe degradations.
Concatenation often helps relative to raw spectra but does not consistently
improve on LVSE alone. The evidence supports keeping local encoding as a
candidate, rather than assuming that a global nonnegative representation adds
useful information in every NIR problem.

\subsection{Classification supports the same practical conclusion}
Across the 76 classification datasets, LVSE increases weighted F1 by
\TwoFiveFOneGain, \ThreeFOneGain\ and \ThreeFiveFOneGain\ pp on average
(Figure~\ref{fig:representations}b,d; Table~\ref{tab:classification}). The metadata-labelled NIR mean gains are
\TwoFiveNIRFOneGain, \ThreeNIRFOneGain\ and \ThreeFiveNIRFOneGain\ pp.
Medians are smaller than means, so the average gain should not be read as the
expected improvement for every new dataset. GCU alone again has a negative
average effect. Concatenation has a slightly larger mean than LVSE for some
versions, without a uniform advantage across datasets.
The steep part of the classification curves lies close to zero: many tasks
change little, while a smaller set gains several percentage points. The
dataset-level maps in Appendix~\ref{app:catalogue} show both this heterogeneity
and the contrasting behaviour of the global branch. The complete catalogue
performance map in Supplementary Figure~\ref{fig:catalogcomplete} retains
every recorded procedure on both complete evaluated panels; its median
absolute-score aggregation differs from these mean paired gains.

The 15-dataset completion campaign contains 180 valid evaluations, no missing
keys and no duplicate keys. The resulting classification score matrix has 76
rows and 18 method columns with no missing cells. This is completeness for the
evaluated classification panel, not evidence that every catalogue entry has a
usable classification target. The ManyClass results also should not be labelled
as native 2.5 inference.

\subsection{The representation also helps PLS and Ridge}
On the same 61 numerical representatives, LVSE improves PLS on \PLSWins/61
and Ridge on \RidgeWins/61 (Table~\ref{tab:matchednumeric}).
Median RMSE reductions are \PLSMedianGain\% and
\RidgeMedianGain\%, respectively. The Ridge mean is only \RidgeMeanGain\%,
showing that local encoding can still produce costly failures despite a positive
median. PLS has a clearer average improvement of \PLSMeanGain\%.
These results motivate evaluating the representation in a classical selection
bank; they do not evaluate a combined raw-versus-LVSE selector.
The first two rows of Figure~\ref{fig:representations}a place the classical
models beside the foundation-model checkpoints. This is the link from a
Raman-inspired input transformation to a general calibration component.

The behavior of GCU mirrors the TabPFN results: standalone global compression
has negative means and medians for both classical models. PLS benefits from
some concatenated representations, whereas Ridge is less consistent. Model
selection must therefore distinguish a useful LVSE branch from the broader
claim that every Raman-inspired representation will transfer to NIR.

\subsection{An AOM extension is feasible, but its marginal gain is small}
\begin{figure}[htbp]\centering
\includegraphics[width=\linewidth]{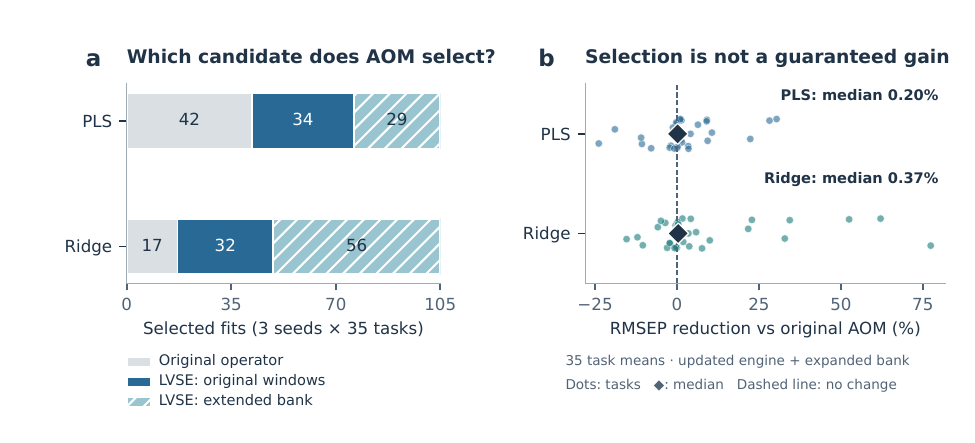}
\caption{AOM uses LVSE candidates frequently, while the aggregate marginal gain
remains small. (a) Training-selected candidate families in 105 depth-one fits per
model (35 tasks, three seeds). Numbers are counts; hatching marks the extended
LVSE bank. (b) Paired task-level RMSEP reductions against archived AOM, averaged
over the three seeds. Each dot is a task; diamonds mark medians. Points are
offset vertically only for visibility. Bank expansion and numerical-engine
changes occurred together, so this is an implementation-and-bank comparison.}
\label{fig:aom}
\end{figure}

The separate original-bank-versus-extended AOM comparison gives median RMSEP
reductions of 0.20\% for depth-one AOM-PLS and 0.37\% for AOM-Ridge
(Figure~\ref{fig:aom}; Table~\ref{tab:aom}). These are much smaller than the LVSE-versus-raw TabPFN
gains. At deeper PLS chain depths, median gains are numerically zero even though
some tasks improve. FastAOM also has approximately zero median gain and loses
on multiple tasks. An enlarged bank can displace a previously useful screened
candidate, so adding a candidate is not guaranteed to improve the final
held-out prediction.

The depth-one campaigns selected an LVSE branch in \AOMPLSLVSESelections/105 AOM-PLS fits and
\AOMRidgeLVSESelections/105 AOM-Ridge fits. This establishes that the candidates are used by training
selection, not that their isolated causal effect is positive. The adaptive
extension has approximately zero median gain against the archive and does not
improve the fixed variant's aggregate median. Historical fitting cost is also
substantially larger after the bank expansion. Together with the changed PLS
fallback, these results support a cautious integration decision: make LVSE
available, retain the original branches, and avoid advertising a general AOM
accuracy or speed improvement from this comparison.

\FloatBarrier

\subsection{A common test panel puts AOM and TabPFN side by side}
\begin{figure}[htbp]\centering
\includegraphics[width=\linewidth]{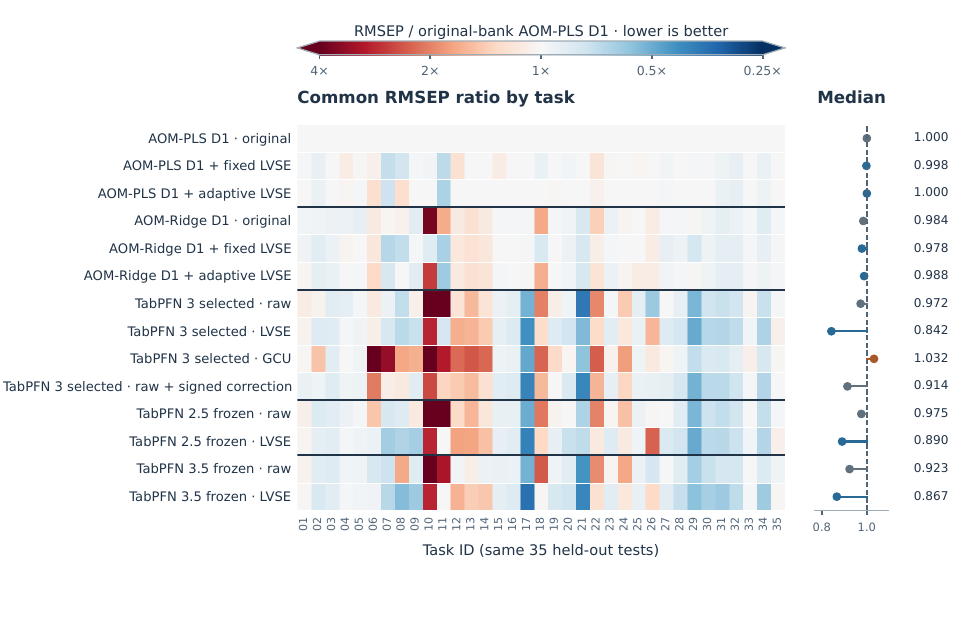}
\caption{A common-test view of AOM and TabPFN. Fourteen explicitly named
pipelines are compared on the same 35 tests using the archived AOM-PLS depth-one
RMSEP as reference on each task. Blue denotes lower error and red higher
error; the colour scale is logarithmic and saturates outside 0.25--4.
Right-hand dots and numbers are medians of the 35 task ratios. Fixed AOM-Ridge
has ratio 0.978 and the selected LVSE-TabPFN pipeline 0.842. ``Selected'' denotes
TabPFN 3 with training-selected representations; ``frozen'' denotes checkpoint
evaluation using those retained representation choices. Appendix~\ref{app:fullscores} contains all 46
pipelines and the task-ID key. Sample caps and selection budgets differ across
model families.}
\label{fig:common}
\end{figure}

Figure~\ref{fig:common} reports saved scores using one common reference,
the retained original-bank AOM-PLS depth-one RMSEP on each of the 35 tasks. The extended
AOM-Ridge fixed variant has a median ratio of 0.978, while the training-selected
TabPFN-LVSE pipeline has a ratio of 0.842. This shows competitive prediction from
both families, with a stronger descriptive score for the LVSE-TabPFN pipeline on
this panel. It does not show that TabPFN has a universally lower cost or that
the ratio is an independent estimate on unseen datasets.
The task-wise map also reveals that the ranking varies with the dataset.
Local encoding often changes the TabPFN row substantially, while the AOM
variants tend to remain closer to their archived counterparts. The right-hand
summary compresses that map to a median without choosing a model on each test.
Exact summary values are retained in Table~\ref{tab:common}.
Supplementary Figure~\ref{fig:suppscorecard} puts the labelled median gains,
common-reference scores and task win/tie/loss bars side by side for all 46
pipelines. It makes the distinction between a representation gain and an
absolute common-test ranking explicit. Supplementary
Figure~\ref{fig:supppoints} retains every normalized task score as a point;
the spread shows why an aggregate ranking cannot determine the best pipeline
for a new task.

\FloatBarrier

The common-panel checkpoint comparison reuses frozen representation choices
from the initial TabPFN-3 evaluation; it does not reselect a new bank for every checkpoint as the
catalogue version comparison does. Its sample caps also differ between AOM and
TabPFN. We therefore use it as a matched-test score comparison, not as a
controlled comparison of identical training budgets. The minimum score in a
test row is descriptive and is never used to define a deployable selector.

\section{Discussion}
The main result is simple: local spectral coordinates can improve a tabular
foundation model on spectral inputs, and their value is not restricted to that model.
LVSE supplies several directions within each neighbourhood instead of asking
the predictor to read every correlated channel directly. This is consistent
with, but does not prove, a useful denoising or regularization effect. PLS and
Ridge results show that the representation is a reusable calibration component,
not necessarily a property of foundation-model weights alone. Our experiment
does not isolate locality itself: scaling, optional SNV and dimensional reduction
are bundled in the procedure, and SNV-only, global-PCA and matched-dimension
controls are absent. Local spectral representations also fit a longer
chemometric tradition of local interval models \citep{norgaard2000}.
The visual evidence separates two practical lessons. Figure~\ref{fig:representations}
shows the benefit of changing the input coordinates, including for classical
models. Figure~\ref{fig:aom} shows that a frequently selected extra component
can still add little to an already flexible procedure. Figure~\ref{fig:common}
places both observations in the context of common held-out tests, where the
preferred pipeline depends on the task.

The global branch provides an equally useful negative result. A global
nonnegative factorization with our clipping, ranks and iteration budgets does
not reliably transfer the benefits observed in the motivating Raman setting.
That observation concerns the tested implementation and these datasets; it
does not invalidate global unmixing in general. Differences in spectral band
structure, normalization, acquisition and optimization may all matter.

The results motivate exposing raw and LVSE branches in nirs4all's pipeline
selection tools \citep{nirs4all2026}. A procedure choosing between these branches
must itself be assessed on held-out data; the present catalogue comparison
selects configurations within each branch. Catalogue loading and native AOM
fits use components of the nirs4all ecosystem, with experiment-specific
orchestration. A single public-pipeline execution route and equivalence between
orchestration routes were not evaluated here.

There are several limits to the conclusion. These are exploratory
campaigns on overlapping catalogue sources, with within-dataset splits and
branch-specific selection budgets. The benchmark does not establish robustness
to new instruments, collection campaigns or extrapolated concentrations.
Channel-based windows are not physically normalized across instruments.
The classification labels and unsupported-class filtering change the task
population. Weighted F1 does not assess probability calibration. Multiple
candidate banks and model versions were inspected, and the AOM update contains
a numerical-engine change. Finally, the ManyClass wrapper adds an inference
procedure that must be identified in a checkpoint comparison. These limits
argue for reporting paired scores and denominators rather than a universal
ranking or a single claim of state-of-the-art performance.

\section{Conclusion}
The results support LVSE as a useful representation candidate for
spectral prediction. On the task-matched numerical subset it improves raw-spectrum TabPFN on average across three
checkpoints, and gives positive median gains for both PLS and Ridge. GCU alone
is less reliable, and combining global and local coordinates is not invariably
better than local encoding. LVSE can be included in AOM's search space, but the
current extension comparison establishes feasibility and task-dependent gains,
not a general improvement of AOM. The appropriate recommendation is to make
LVSE available in the nirs4all ecosystem and evaluate training-only selection
alongside raw spectra and established preprocessing alternatives. The full
catalogue overview remains exploratory because coverage, target semantics and
spectral-family annotations require care.

\section*{CRediT authorship contribution statement}
Gr\'egory Beurier: Conceptualization, Methodology, Software, Formal analysis,
Investigation, Data curation, Validation, Visualization, Writing -- original
draft, Project administration. Robin Reiter: Methodology, Software, Validation,
Writing -- review \& editing. Denis Cornet: Methodology, Resources, Supervision,
Validation, Writing -- review \& editing. Lauriane Rouan: Conceptualization,
Methodology, Validation, Supervision, Writing -- review \& editing.

\section*{Declaration of competing interest}
The authors declare that the software described in this work is distributed
under open-source licences and that optional commercial licences for closed-source
or software-as-a-service use are available through CIRAD; commercial-licence
enquiries are directed to \texttt{nirs4all-admin@cirad.fr}. The authors declare
no other known competing financial interests or personal relationships that
could have appeared to influence the work reported in this paper.

\section*{Funding}
This research received no specific grant from any funding agency in the public,
commercial or not-for-profit sectors.

\section*{Data and code availability}
Dataset source cards and original retrieval and licence information are
maintained at \url{https://datasets.nirs4all.org} and
\url{https://github.com/GBeurier/nirs4all-datasets}. The public software entry
points are \url{https://github.com/GBeurier/nirs4all} and
\url{https://github.com/GBeurier/nirs4all-methods}. The research implementations
and frozen partitions used for these experiments are maintained by the authors;
this draft does not claim that all experimental scripts are already public.
The accompanying evidence bundle contains paired seed scores, dataset
aggregates, absolute common-test scores, protocol snapshots and source hashes.
Its aggregation script regenerates the manuscript tables without fitting any
models. Underlying spectra are not redistributed, and their original terms
remain applicable. A public archive identifier for this specific evidence bundle
has not yet been assigned.

\section*{Declaration of Generative AI and AI-assisted technologies in the writing process}
During preparation of this manuscript, Anthropic Claude and OpenAI tools were
used to assist with code review, implementation checks, repository organization,
benchmark aggregation scripts, LaTeX editing and language revision. The authors
retain full responsibility for the final content of the article. Numerical
results, references, code and manuscript content remain subject to the authors'
final review and verification before submission.

\clearpage
\appendix
\section{Numerical summaries}\label{app:numbers}
The exact summaries underlying the main figures are retained below. The regression
and classification denominators and their aggregation differ as specified in the
Methods; their numerical scales should not be compared directly.

\begin{table}[htbp]\centering\small
\caption{Primary post-hoc numerical-target comparison. All five predictors use
the same 61 representative source--target tasks and three seeds after exclusion
of ten nominal-code, metadata or semantically ambiguous targets. Gains are relative to the
raw input of the same predictor. This is not a certified NIR-only cohort.}
\label{tab:matchednumeric}
\begin{tabular}{llrrrr}
\toprule
Model & Input & $N$ & Mean (\%) & Median (\%) & W/L \\
\midrule
PLS & GCU & 61 & -26.33 & -7.40 & 13/48 \\
PLS & GCU+LVSE & 61 & 2.60 & 4.49 & 42/19 \\
PLS & LVSE & 61 & 5.45 & 4.75 & 47/14 \\
Ridge & GCU & 61 & -38.14 & -12.70 & 6/55 \\
Ridge & GCU+LVSE & 61 & -0.84 & 1.76 & 36/25 \\
Ridge & LVSE & 61 & 0.57 & 1.89 & 36/25 \\
TabPFN v2.5 & GCU & 61 & -32.61 & -6.95 & 13/48 \\
TabPFN v2.5 & GCU+LVSE & 61 & 8.72 & 4.41 & 48/13 \\
TabPFN v2.5 & LVSE & 61 & 10.22 & 4.99 & 47/14 \\
TabPFN v3 & GCU & 61 & -21.37 & -2.97 & 18/43 \\
TabPFN v3 & GCU+LVSE & 61 & 11.41 & 4.99 & 50/11 \\
TabPFN v3 & LVSE & 61 & 12.80 & 4.74 & 49/12 \\
TabPFN v3.5 & GCU & 61 & -28.83 & -2.15 & 23/38 \\
TabPFN v3.5 & GCU+LVSE & 61 & 13.05 & 6.84 & 52/9 \\
TabPFN v3.5 & LVSE & 61 & 14.52 & 7.82 & 53/8 \\
\bottomrule
\end{tabular}

\end{table}

\begin{table}[htbp]\centering\small
\caption{Paired regression gains relative to raw spectra within each TabPFN
version, on the full 71-entry declared-numeric panel, including numerical codes
and metadata targets. W/L counts dataset wins/losses; there are no ties.
GCU+LVSE means concatenated coordinates.}\label{tab:regression}
\begin{tabular}{llrrrr}
\toprule
TabPFN & Input & $N$ & Mean (\%) & Median (\%) & W/L \\
\midrule
v2.5 & GCU & 71 & -46.98 & -7.55 & 14/57 \\
v2.5 & LVSE & 71 & 8.68 & 4.86 & 53/18 \\
v2.5 & GCU+LVSE & 71 & 7.76 & 3.40 & 53/18 \\
v3 & GCU & 71 & -35.67 & -3.82 & 18/53 \\
v3 & LVSE & 71 & 3.86 & 4.72 & 54/17 \\
v3 & GCU+LVSE & 71 & 6.55 & 4.11 & 55/16 \\
v3.5 & GCU & 71 & -41.41 & -2.63 & 26/45 \\
v3.5 & LVSE & 71 & 11.88 & 8.26 & 61/10 \\
v3.5 & GCU+LVSE & 71 & 7.68 & 7.15 & 60/11 \\
\bottomrule
\end{tabular}

\end{table}

\begin{table}[htbp]\centering\small
\caption{Weighted-F1 changes relative to raw spectra. All-dataset means and
medians use 76 datasets; the metadata-labelled NIR stratum uses 55. W/L excludes ties, whose counts can
be recovered from $76-$W$-$L. Version 2.5 includes 15 ManyClass datasets;
Appendix~\ref{app:manyclass} separates them.}\label{tab:classification}
\begin{tabular}{llrrrr}
\toprule
TabPFN & Input & Mean (pp) & Median (pp) & NIR-label mean (pp) & W/L \\
\midrule
v2.5 & GCU & -3.22 & -1.85 & -3.72 & 11/61 \\
v2.5 & LVSE & 2.22 & 0.63 & 2.71 & 54/12 \\
v2.5 & GCU+LVSE & 2.49 & 0.67 & 2.65 & 53/12 \\
v3 & GCU & -2.98 & -1.41 & -3.39 & 15/58 \\
v3 & LVSE & 2.56 & 0.68 & 2.52 & 52/15 \\
v3 & GCU+LVSE & 2.57 & 0.66 & 2.40 & 57/8 \\
v3.5 & GCU & -1.07 & -0.39 & -1.63 & 20/46 \\
v3.5 & LVSE & 3.81 & 1.43 & 3.77 & 57/7 \\
v3.5 & GCU+LVSE & 3.87 & 1.17 & 3.78 & 58/7 \\
\bottomrule
\end{tabular}

\end{table}

\begin{table}[htbp]\centering\small
\caption{Original all-target classical comparison: 966 declared numerical
source--target pairs aggregated to 71 catalogue identifiers. Each procedure
is compared with its own training-selected raw baseline. This target population
differs from the matched 61-task analysis.}\label{tab:classical}
\begin{tabular}{llrrrr}
\toprule
Model & Input & $N$ & Mean (\%) & Median (\%) & W/L \\
\midrule
PLS & GCU & 71 & -25.17 & -8.13 & 8/63 \\
PLS & LVSE & 71 & 3.63 & 3.70 & 52/19 \\
PLS & GCU+LVSE & 71 & -3.08 & 3.61 & 49/22 \\
Ridge & GCU & 71 & -36.23 & -12.41 & 5/66 \\
Ridge & LVSE & 71 & 0.15 & 1.17 & 44/27 \\
Ridge & GCU+LVSE & 71 & -4.57 & -0.10 & 35/36 \\
\bottomrule
\end{tabular}

\end{table}

\begin{table}[htbp]\centering\small
\caption{Selected, explicitly named pipelines from the completed 46-pipeline
common-test comparison. All use the same 35 tasks derived from the AOM benchmark
\citep{beurier2026aom}. Ratios use retained original-bank AOM-PLS depth one.
Selected/frozen representation procedures are distinguished;
the full 46-pipeline table and absolute RMSEP values accompany the source data.}
\label{tab:common}
\begin{tabular}{lr}
\toprule
Pipeline & Median RMSEP ratio \\
\midrule
AOM-PLS D1 -- original & 1.000 \\
AOM-PLS D1 + fixed LVSE & 0.998 \\
AOM-PLS D1 + adaptive LVSE & 1.000 \\
AOM-Ridge D1 -- original & 0.984 \\
AOM-Ridge D1 + fixed LVSE & 0.978 \\
AOM-Ridge D1 + adaptive LVSE & 0.988 \\
TabPFN 3 selected -- raw & 0.972 \\
TabPFN 3 selected -- LVSE & 0.842 \\
TabPFN 3 selected -- GCU & 1.032 \\
TabPFN 3 selected -- raw + signed correction & 0.914 \\
TabPFN 2.5 frozen -- raw & 0.975 \\
TabPFN 2.5 frozen -- LVSE & 0.890 \\
TabPFN 3.5 frozen -- raw & 0.923 \\
TabPFN 3.5 frozen -- LVSE & 0.867 \\
\bottomrule
\end{tabular}

\end{table}
\FloatBarrier

\begin{table}[htbp]\centering\small
\caption{Updated fixed-LVSE bank versus archived AOM on 35 tasks. W/T/L gives
task wins/ties/losses under the campaign's existing tolerance. Historical fit
ratios are updated/archived times, so values above one mean more time; the
campaigns are not a controlled timing experiment.}\label{tab:aom}
\begin{tabular}{lrrrr}
\toprule
Model & Depth & Median (\%) & W/T/L & Historical fit ratio \\
\midrule
AOM-PLS & 1 & 0.20 & 18/6/11 & 14.7 \\
AOM-PLS & 2 & 0.00 & 12/16/7 & 15.9 \\
AOM-PLS & 3 & 0.00 & 14/15/6 & 12.6 \\
AOM-Ridge & 1 & 0.37 & 19/4/12 & 8.5 \\
FastAOM-PLS & 1 & 0.00 & 9/12/14 & 22.3 \\
FastAOM-PLS & 2 & 0.00 & 8/15/12 & 59.0 \\
FastAOM-PLS & 3 & 0.00 & 8/16/11 & 68.1 \\
\bottomrule
\end{tabular}

\end{table}
\FloatBarrier

\section{Native and wrapped TabPFN 2.5 classification}\label{app:manyclass}
Table~\ref{tab:manyclass} separates the saved native and ManyClass evaluations.
The wrapper is used for both raw and transformed inputs on each of the 15
unsupported datasets. It does not change their labels, held-out samples or
outer seeds. Gains are paired within each procedure, so the wrapped raw baseline
is not compared with a native transformed model or vice versa.

\begin{table}[htbp]\centering\small
\caption{Version-2.5 classification changes relative to the corresponding raw
baseline, separated by native and output-coding protocol.}\label{tab:manyclass}
\begin{tabular}{llrrr}
\toprule
TabPFN 2.5 protocol & Input & $N$ & Mean (pp) & Median (pp) \\
\midrule
Native checkpoint & GCU & 61 & -2.33 & -1.44 \\
Native checkpoint & LVSE & 61 & 1.62 & 0.46 \\
Native checkpoint & GCU+LVSE & 61 & 1.90 & 0.46 \\
Checkpoint + ManyClass & GCU & 15 & -6.87 & -3.90 \\
Checkpoint + ManyClass & LVSE & 15 & 4.63 & 1.74 \\
Checkpoint + ManyClass & GCU+LVSE & 15 & 4.88 & 1.92 \\
\bottomrule
\end{tabular}

\end{table}

\section{Reading the accompanying score files}
\texttt{paired\_scores\_by\_seed.csv} records the dataset, spectral source,
target, family, task type, seed, model, representation, paired absolute scores
and gain. \texttt{paired\_gains\_by\_dataset.csv} performs the declared seed/task
averaging, and \texttt{representation\_summary.csv} gives unrounded aggregate
results, win counts and the NIR and ManyClass strata.
\texttt{scores\_with\_provenance.csv} retains the saved method details, including
the selected representation and the output-coding marker where available.
The common-test files retain absolute RMSEP values, the common-reference summary,
the archived-versus-updated AOM comparisons and the adaptive summary. Thus the
full evidence includes unfavorable variants, rather than only the pipelines
highlighted in Table~\ref{tab:common}.

\texttt{provenance.json} identifies the input CSVs and protocol manifests by
SHA-256 and records the aggregation equations. These hashes identify the saved
evidence; they are not proof of numerical equivalence between separate software
execution routes. The paper's tables were generated from the saved scores.
No new model training or inference was performed to prepare this paper.

The primary 61-task analysis is recorded as paired seed scores in
\path{matched_numerical_pairs.csv}, dataset gains in
\path{matched_numerical_gains.csv}, and summary statistics in
\path{matched_numerical_summary.csv}. The ten excluded representatives are
listed in \path{regression_task_exclusions.csv}.
The complete performance matrices are retained in
\path{catalogue_regression_complete.csv} and
\path{catalogue_classification_complete.csv}, with dataset and method
label keys. These are unchanged snapshots of the saved matrices, preserving
their row and column order. They contain 2,201 numerical and 1,368 classification
scores, with no missing cells. They summarize available task/seed scores by
medians, rather than using the primary analysis's mean paired gains.

\FloatBarrier
\section{Complete common-test comparison}\label{app:fullscores}
\begin{figure}[p]\centering
\includegraphics[width=\linewidth,height=0.81\textheight,keepaspectratio]{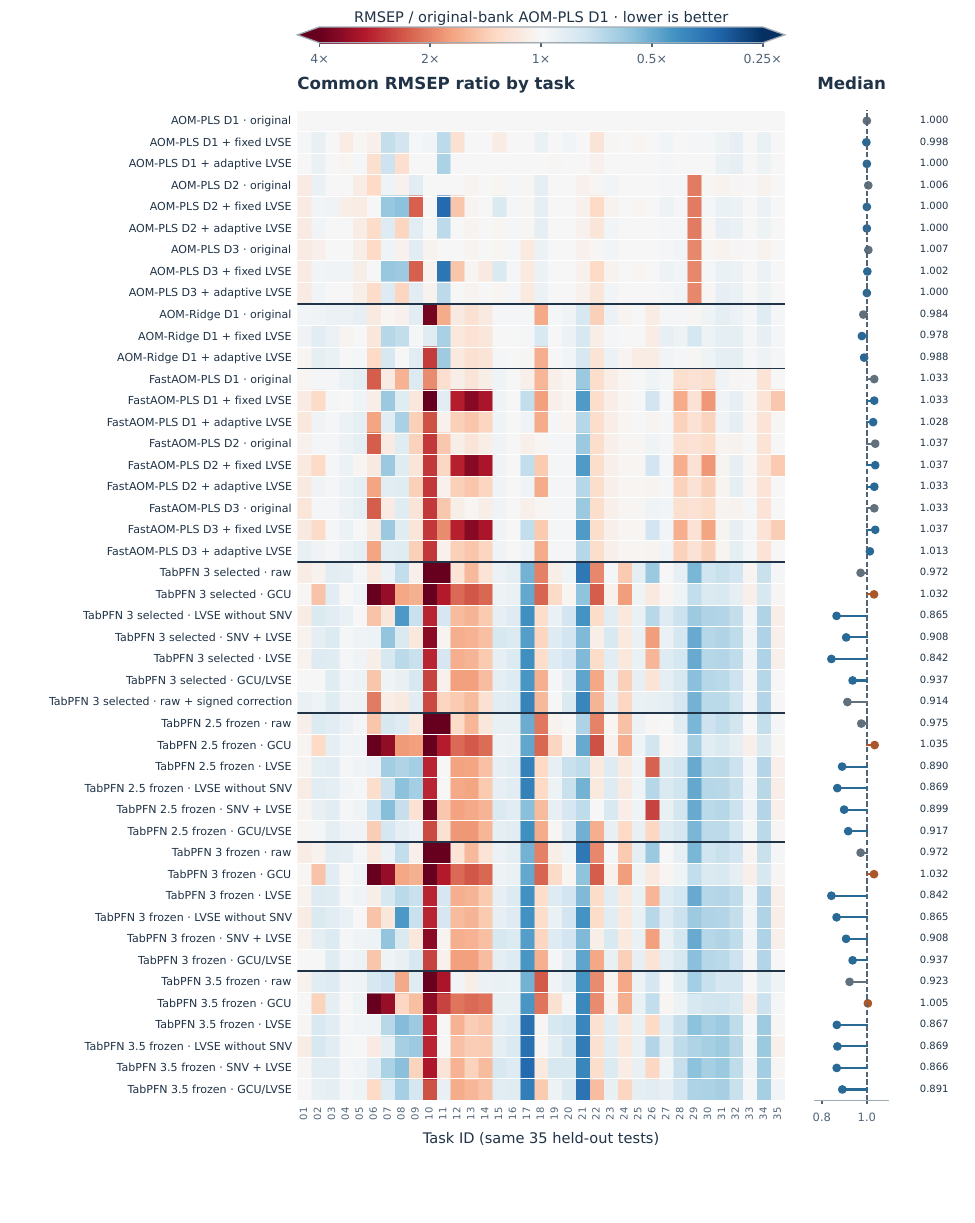}
\caption{All 46 pipelines on the 35 tests derived from the AOM benchmark
\citep{beurier2026aom}. Ratios, colours and medians
have the same definitions as Figure~\ref{fig:common}. Colour saturation outside
0.25--4 does not discard scores. Multiple rows are near the reference; LVSE
helps the TabPFN variants on many tasks, but the task-wise pattern is
heterogeneous. The initial and version-specific TabPFN campaigns retain their
distinct selection procedures.}
\label{fig:complete}
\end{figure}

Figure~\ref{fig:complete} retains every scored pipeline on the same common
reference, including the less successful global-coordinate and FastAOM
variants. The numerical matrix accompanies the evidence bundle.
Table~\ref{tab:taskkey} maps the task IDs used in both common-test figures.

\FloatBarrier
\begin{table}[htbp]\centering\scriptsize
\caption{Task-ID key for Figures~\ref{fig:common}, \ref{fig:complete},
and \ref{fig:supppoints}.
Identifiers preserve the campaign's partition and instrument labels.}
\label{tab:taskkey}
\begin{tabular}{rl}
\toprule
ID & Task \\
\midrule
01 & \texttt{An\_spxyG70\_30\_byCultivar\_MicroNIR} \\
02 & \texttt{TIC\_spxy70} \\
03 & \texttt{An\_spxyG70\_30\_byCultivar\_NeoSpectra} \\
04 & \texttt{WUEinst\_spxyG70\_30\_byCultivar\_MicroNIR\_NeoSpectra} \\
05 & \texttt{An\_spxyG70\_30\_byCultivar\_MicroNIR\_NeoSpectra} \\
06 & \texttt{Beer\_OriginalExtract\_60\_KS} \\
07 & \texttt{Beer\_OriginalExtract\_60\_YbaseSplit} \\
08 & \texttt{Biscuit\_Fat\_40\_RandomSplit} \\
09 & \texttt{Biscuit\_Sucrose\_40\_RandomSplit} \\
10 & \texttt{Corn\_Oil\_80\_ZhengChenPelegYbaseSplit} \\
11 & \texttt{Corn\_Starch\_80\_ZhengChenPelegYbaseSplit} \\
12 & \texttt{DIESEL\_bp50\_246\_b-a} \\
13 & \texttt{DIESEL\_bp50\_246\_hla-b} \\
14 & \texttt{DIESEL\_bp50\_246\_hlb-a} \\
15 & \texttt{An\_spxyG70\_30\_byCultivar\_ASD} \\
16 & \texttt{Beef\_Marbling\_RandomSplit} \\
17 & \texttt{Ccar\_spxyG\_block2deg} \\
18 & \texttt{Rice\_Amylose\_313\_YbasedSplit} \\
19 & \texttt{grapevine\_chloride\_556\_KS} \\
20 & \texttt{ALPINE\_P\_291\_KS} \\
21 & \texttt{Chla+b\_spxyG\_block2deg} \\
22 & \texttt{Rd25\_GTtestSite} \\
23 & \texttt{Rd25\_XSBNtestSite} \\
24 & \texttt{Rd25\_CBtestSite} \\
25 & \texttt{Rd25\_spxy70} \\
26 & \texttt{Chla+b\_spxyG\_species} \\
27 & \texttt{Fv\_Fm\_grp70\_30} \\
28 & \texttt{N\_woOutlier} \\
29 & \texttt{C\_woOutlier} \\
30 & \texttt{N\_wOutlier} \\
31 & \texttt{ph\_groupSampleID\_stratDateVar\_balRows} \\
32 & \texttt{ta\_groupSampleID\_stratDateVar\_balRows} \\
33 & \texttt{LMA\_spxyG70\_30\_byCultivar\_ASD} \\
34 & \texttt{brix\_groupSampleID\_stratDateVar\_balRows} \\
35 & \texttt{LMA\_spxyG\_block2deg} \\
\bottomrule
\end{tabular}

\end{table}
\FloatBarrier

\section{Dataset-level representation gains}\label{app:catalogue}
\begin{figure}[p]\centering
\includegraphics[width=\linewidth,height=0.81\textheight,keepaspectratio]{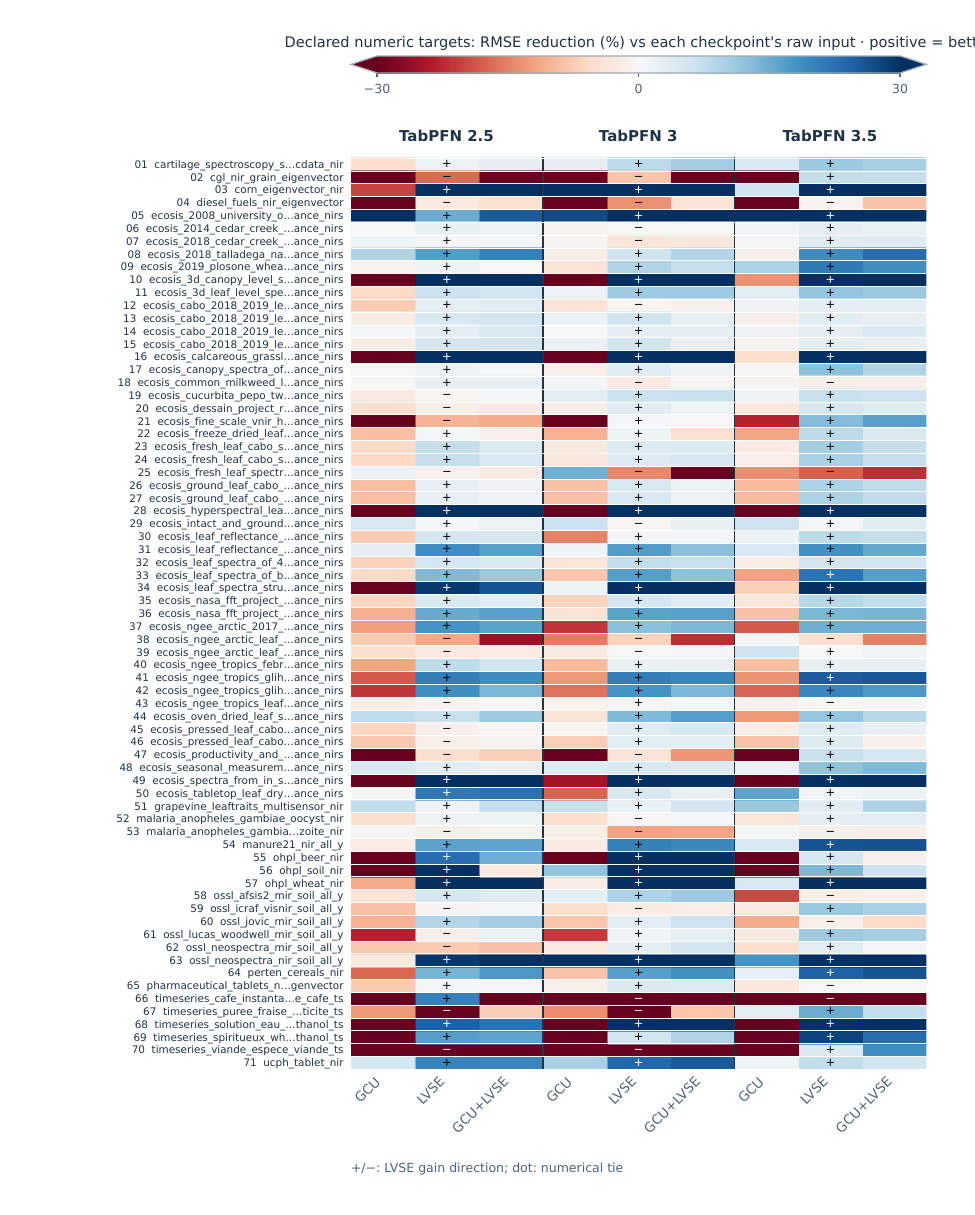}
\caption{Paired gains on the complete 71-entry declared-numeric panel, including
metadata and nominal-code targets. Each cell is the mean
paired RMSE reduction relative to the raw input of the same checkpoint, with
three seeds. Colours saturate at $\pm30\%$; values beyond that range remain in the
source data. Plus/minus marks encode LVSE's
gain direction independently of colour; a dot marks a numerical tie. The
dataset-level heterogeneity motivates further selection experiments. Large
class-code RMSE losses are not evidence of concentration-calibration failure.}
\label{fig:catalogreg}
\end{figure}

\begin{figure}[p]\centering
\includegraphics[width=\linewidth,height=0.81\textheight,keepaspectratio]{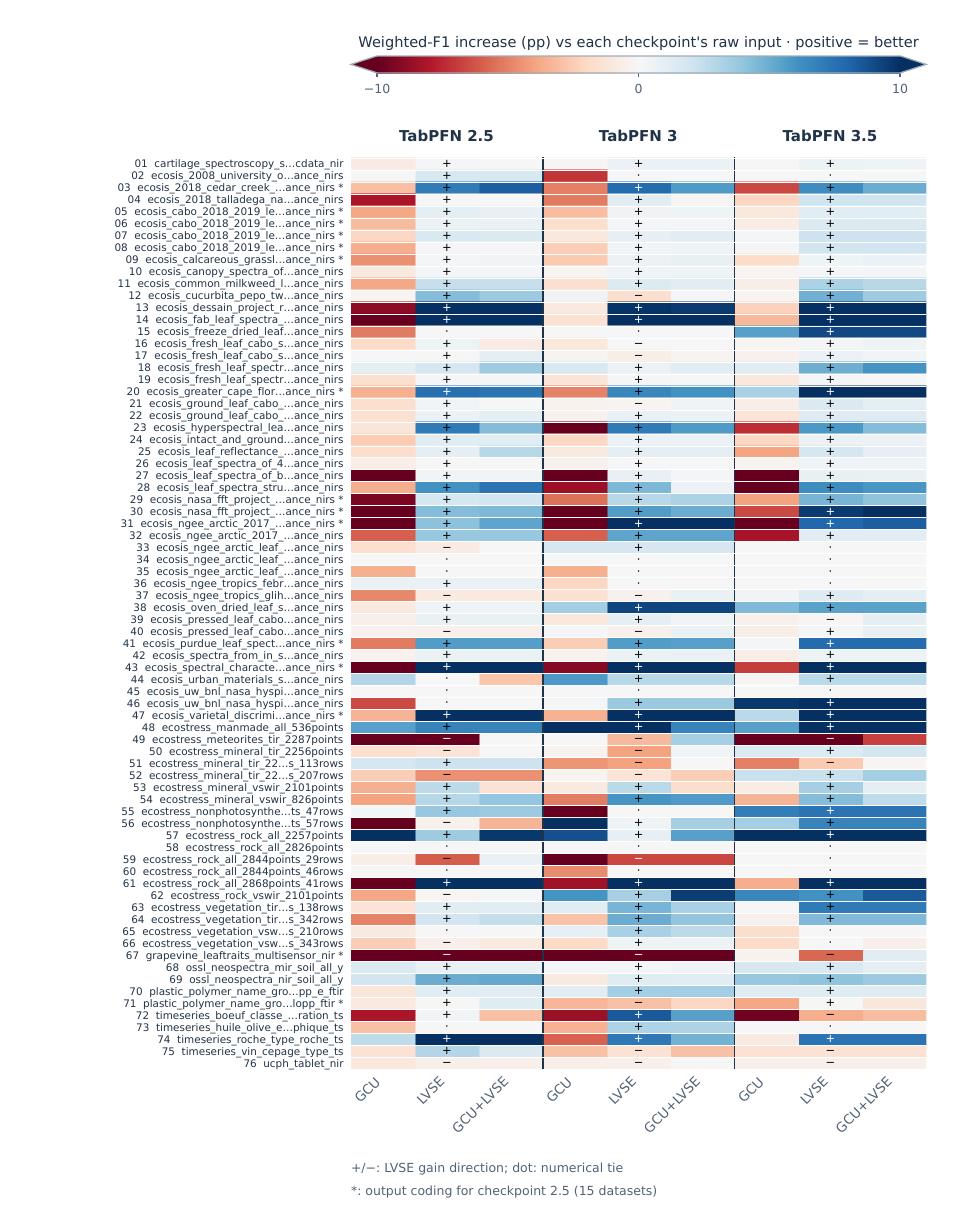}
\caption{Paired weighted-F1 increases on all 76 classification datasets, in
percentage points. Colours saturate at $\pm10$ pp, retaining exact values in
the source data. LVSE signs are marked as in Figure~\ref{fig:catalogreg}.
Asterisks identify the 15 datasets using output coding for checkpoint 2.5;
both raw and represented inputs use the same wrapper within those rows.
Every representation/version comparison is present.}
\label{fig:catalogclass}
\end{figure}

Figures~\ref{fig:catalogreg} and \ref{fig:catalogclass} retain the paired
dataset-level pattern behind the summary figure. Rows are catalogue identifiers,
not independent acquisition cohorts. Each row has all nine representation/version
comparisons. Positive and negative LVSE directions are marked explicitly; exact
values and full dataset names accompany the score bundle. The maps show broad
local-encoding benefits alongside exceptions and a less reliable global branch.

\FloatBarrier

\section{Transform and inference details}\label{app:implementation}
GCU uses coordinate-descent NMF with NNDSVDa initialization, Frobenius
reconstruction loss, tolerance $10^{-3}$ and no explicit regularization. Its
scale is the scalar standard deviation of the capped training matrix. Rank is
clipped to $\min(k,\min(n_{\rm fit},p)-1)$. New-row encoding holds the basis
fixed and optimizes nonnegative coordinates with the same iteration limit and
tolerance. Convergence counts were not recorded, so convergence cannot be
retrospectively certified for every fit.

LVSE floors normalization denominators at $10^{-8}$, retains final shorter
windows, and clips rank to one less than the minimum of fitting rows and window
width. A one-channel final window is skipped. Zero-variance channels become
zero after scaling; there is no singular-value threshold or component whitening.
The fixed AOM extension adds window/rank pairs $(32,4)$ and $(256,8)$, and
50\%-overlap pairs $(64,8)$ and $(128,4)$, to the four nonoverlapping
$(64,2)$, $(64,4)$, $(128,2)$ and $(128,4)$ candidates.

The versioned classifier/regressor checkpoints follow these filename patterns,
where the task field is either classifier or regressor:
\begin{quote}\small
\texttt{tabpfn-v2.5-<task>-v2.5\_default.ckpt}\\
\texttt{tabpfn-v3-<task>-v3\_default.ckpt}\\
\texttt{tabpfn-v3.5-20260909.safetensors}
\end{quote}
Version runners use 100-query batches;
the original default runner uses batches of 2,048. Archived checkpoint hashes
and an explicit base-default checkpoint identifier remain unavailable. This
limits exact weight-level reproduction from the score bundle alone.

\clearpage
\section*{Supplementary figures}
\setcounter{figure}{0}
\renewcommand{\thefigure}{S\arabic{figure}}
\renewcommand{\theHfigure}{supp.\arabic{figure}}

\begin{figure}[!htbp]\centering
\includegraphics[width=\linewidth,height=0.76\textheight,keepaspectratio]{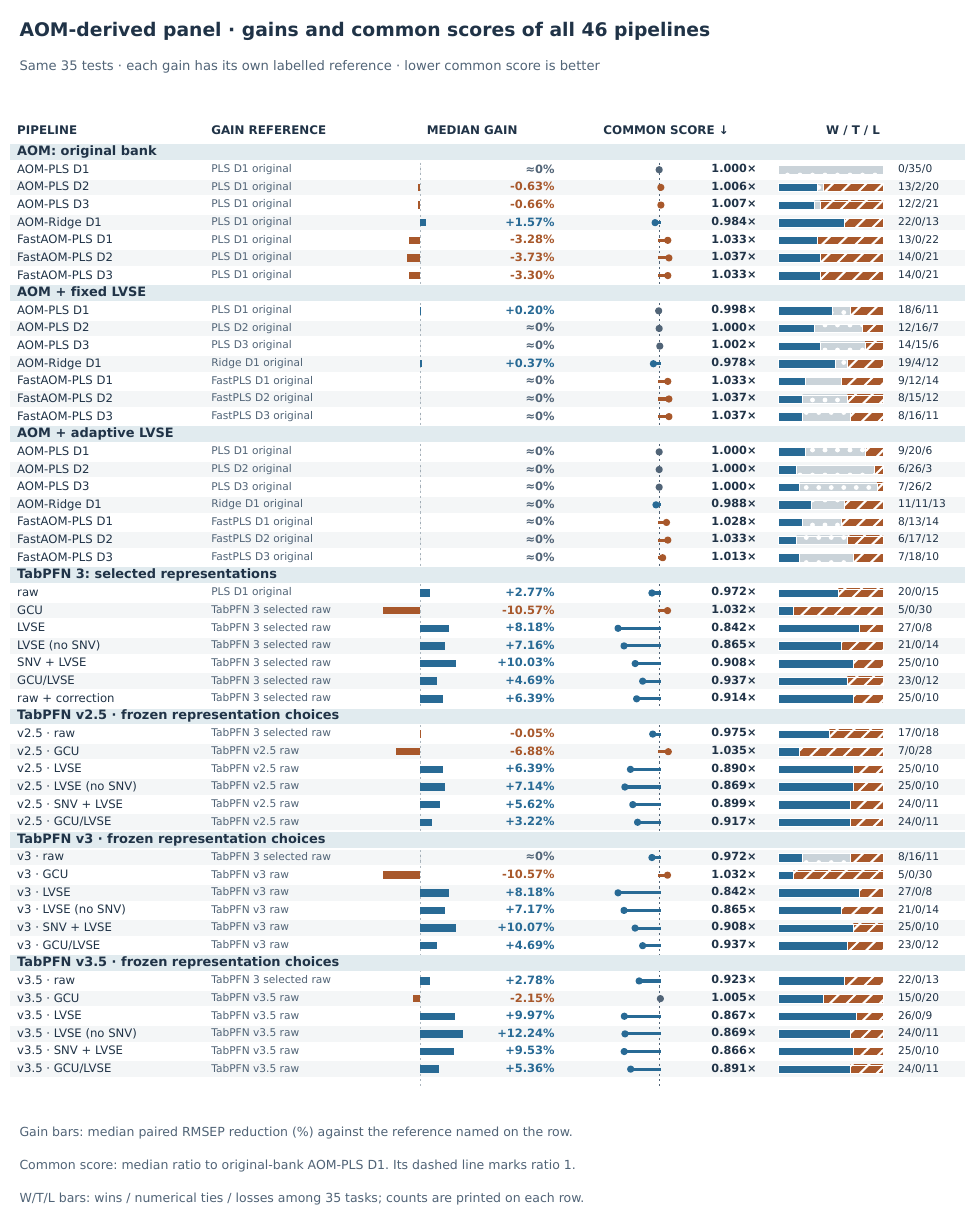}
\caption{Gains, common scores and win/tie/loss bars for all 46 pipelines on
the 35-task AOM-derived panel \citep{beurier2026aom}. Gain bars and percentages
use the reference named on each row; common scores are medians of task-wise
RMSEP ratios to retained original-bank AOM-PLS D1.
W/T/L counts compare the 35 tasks with the row's gain reference. The initial
LVSE-TabPFN gain is 8.18\% versus its raw input, with 27 wins and eight losses;
fixed depth-one AOM median gains are 0.20\% for PLS and 0.37\% for Ridge.}
\label{fig:suppscorecard}
\end{figure}
\FloatBarrier
\clearpage

\begin{figure}[p]\centering
\includegraphics[width=\linewidth,height=0.80\textheight,keepaspectratio]{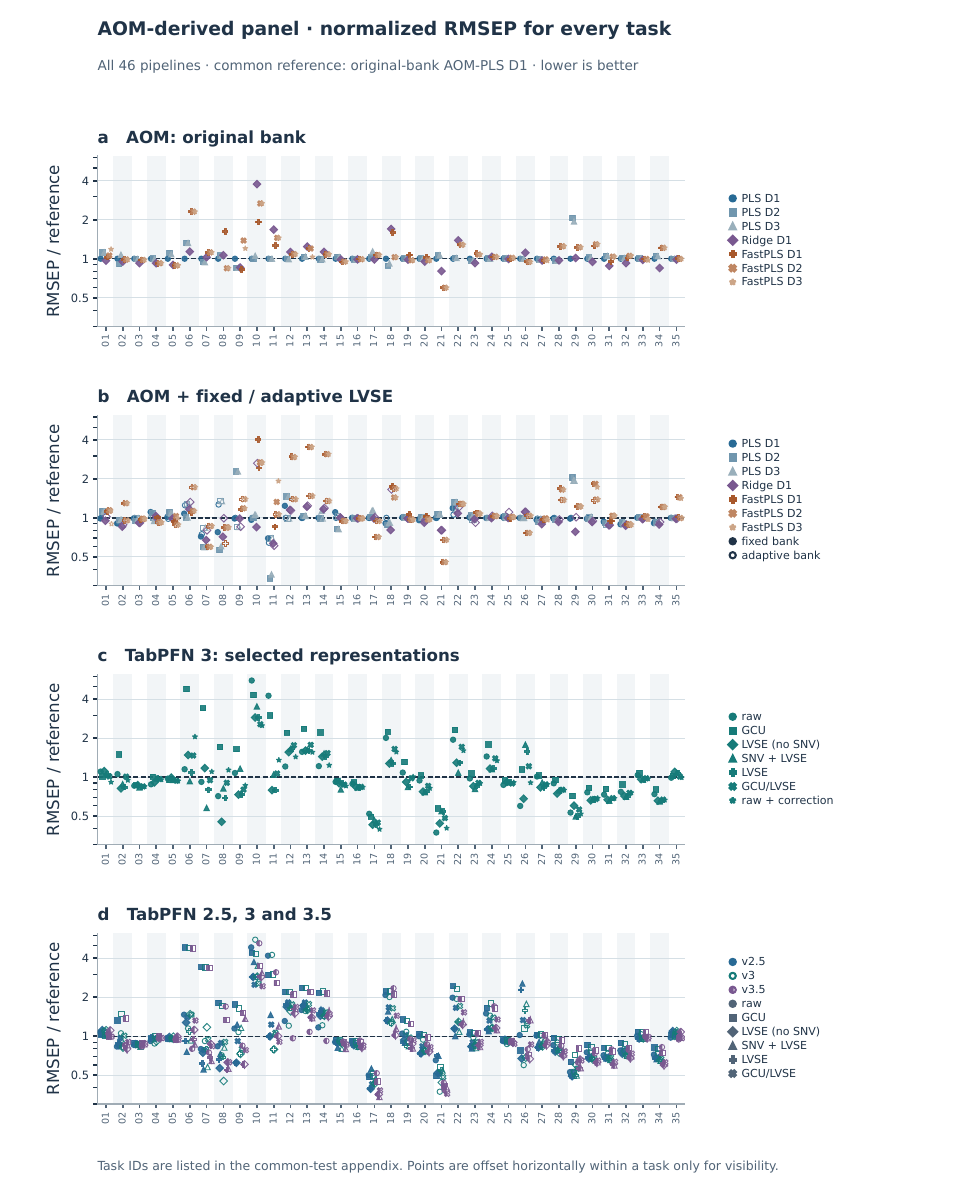}
\caption{Normalized RMSEP by task for all 46 pipelines. Four panels separate
archived AOM, its fixed/adaptive extensions, the initial TabPFN campaign and
three checkpoints. The dashed line at one is archived AOM-PLS D1 on the same
test; the vertical scale is logarithmic. Horizontal offsets only prevent
point overlap. All scores are shown, and task IDs follow Table~\ref{tab:taskkey}.
No uncertainty intervals or test-based winner selection are displayed.}
\label{fig:supppoints}
\end{figure}
\FloatBarrier
\clearpage

\begin{landscape}
\begin{figure}[p]\centering
\includegraphics[width=\linewidth,height=0.82\textwidth,keepaspectratio]{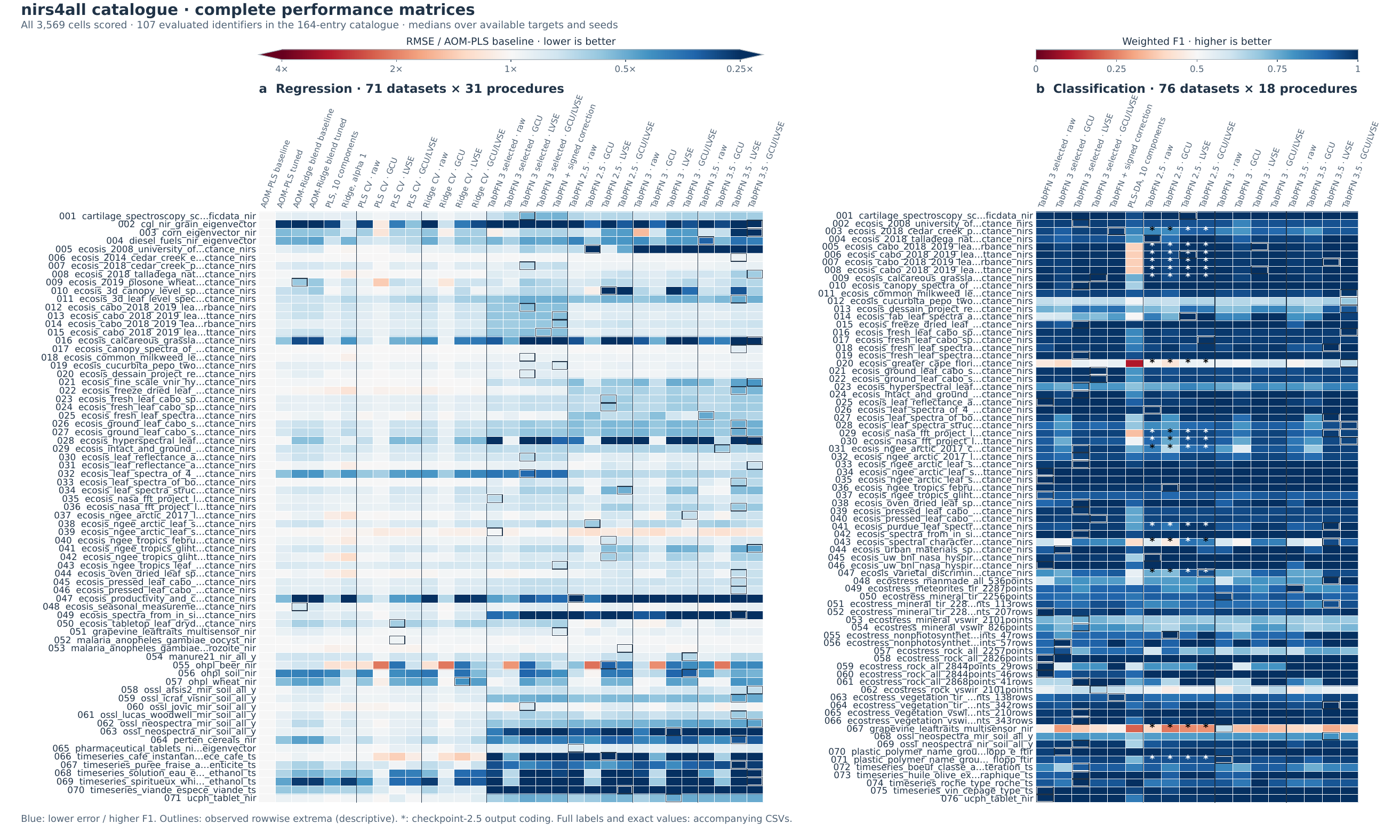}
\caption{Complete saved catalogue matrices: (a) 71 identifiers $\times$ 31
procedures, median task/seed RMSE ratios to AOM-PLS; (b) 76 $\times$ 18,
median weighted F1. All 3,569 cells are scored, in their original independent
row/column orders (107 of 164 catalogue identifiers). Blue indicates lower
error/higher F1; red the converse. Ratio colours saturate outside 0.25--4;
F1 spans 0--1. Outlines mark descriptive rowwise extrema; asterisks mark
checkpoint-2.5 output coding. Columns can summarize different targets;
this exploratory map is distinct from the matched 61-task analysis.
Full labels and values accompany the evidence.}
\label{fig:catalogcomplete}
\end{figure}
\FloatBarrier
\end{landscape}
\end{document}